\documentclass[10pt,twocolumn,letterpaper]{article}

\usepackage[pagenumbers]{cvpr} 

\definecolor{cvprblue}{rgb}{0.21,0.49,0.74}
\usepackage[pagebackref,breaklinks,colorlinks,allcolors=cvprblue]{hyperref}
\usepackage{multirow}
\usepackage{pifont}
\newcommand{\cmark}{\ding{51}} 
\newcommand{\xmark}{\ding{55}} 

\def\paperID{2387} 
\def\confName{CVPR}
\def\confYear{2026}
\usepackage{color}
\definecolor{color1}{rgb}{1.        , 0.62352941, 0.16862745}
\definecolor{color2}{rgb}{0.11764706, 0.27843137, 0.68235294}
\definecolor{color3}{rgb}{0.8627451 , 0.08627451, 0.29019608}
\definecolor{color4}{rgb}{0.31372549, 0.70196078, 0.1372549}
\definecolor{color5}{rgb}{0.46666667, 0.13333333, 0.02352941}
\title{PhysX-Anything: Simulation-Ready Physical 3D Assets from Single Image
}

\author{Ziang Cao$^{1}$, Fangzhou Hong$^1$, Zhaoxi Chen$^1$,
	{Liang Pan}$^2$,  {Ziwei Liu}$^{1}\footnotemark[1]$\\
	$^{1}$S-Lab, Nanyang Technological University \quad $^{2}$Shanghai AI Lab
	\\
	{\tt\small \url{https://physx-anything.github.io}}
}

\def\ourname{PhysX-Anything}

\def\ournewdata{PhysX-Mobility}

\begin{document}

\twocolumn[{%
\renewcommand\twocolumn[1][]{#1}%
\maketitle
\vspace{-20pt}
\includegraphics[width=1\linewidth]{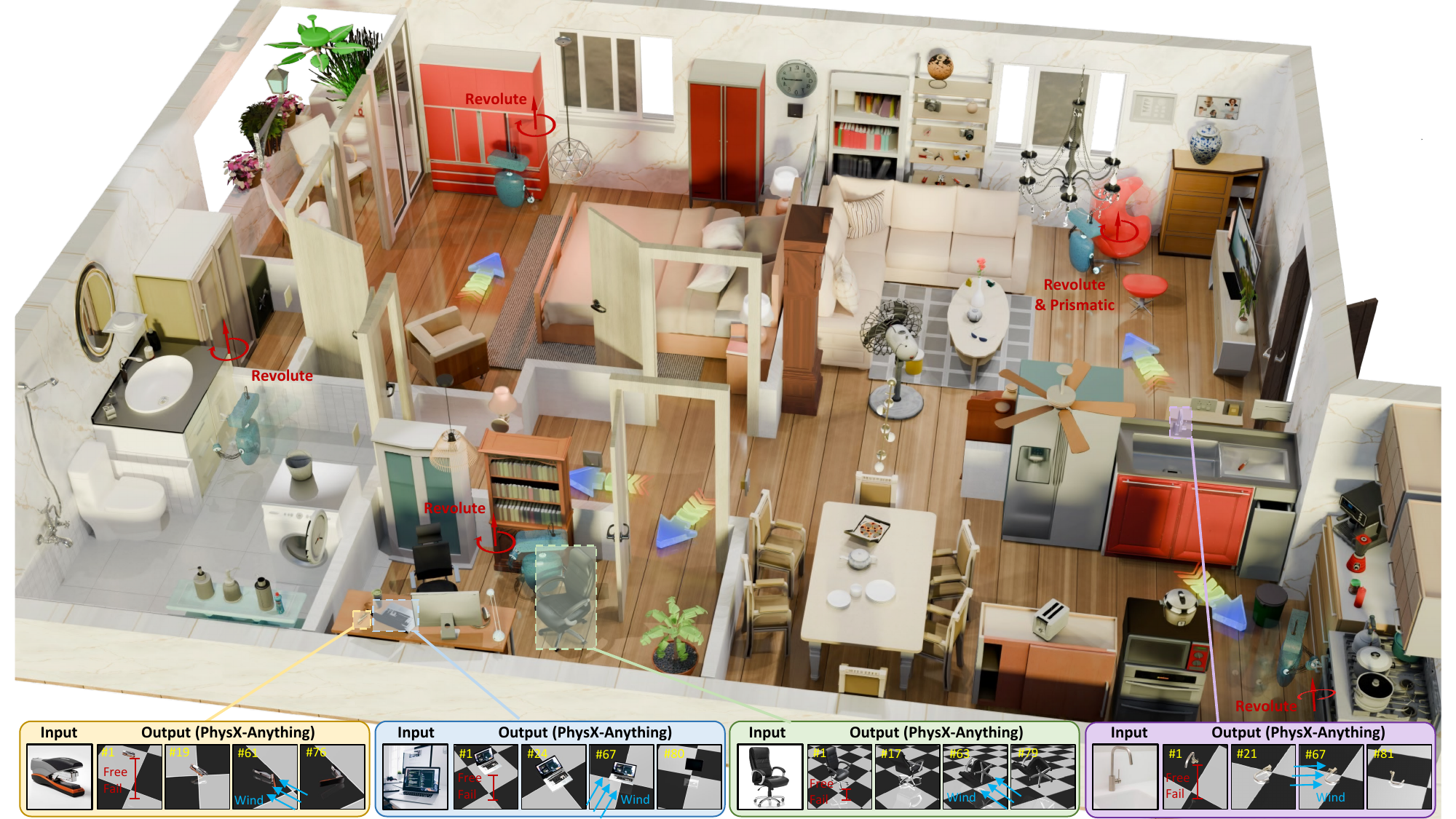}    
\vspace{-15pt}
\captionof{figure}{Given a single real-world image, \ourname\ generates a detailed physical 3D object, recovering both its \textbf{articulation structure and physical properties}, and exports URDF and XML files that can be directly deployed in physics engines. \vspace{20pt}}
\label{fig:teaser}
}]

\begin{abstract}

3D modeling is shifting from static visual representations toward physical, articulated assets that can be directly used in simulation and interaction.
However, most existing 3D generation methods overlook key physical and articulation properties, thereby limiting their utility in embodied AI. To bridge this gap, we introduce \textbf{\ourname}, the first \textbf{simulation-ready} physical 3D generative framework that, given a single in-the-wild image, produces high-quality sim-ready 3D assets with explicit geometry, articulation, and physical attributes. Specifically, we propose the first VLM-based physical 3D generative model, along with a new 3D representation that efficiently tokenizes geometry. It reduces the number of tokens by \textbf{193$\times$}, enabling explicit geometry learning within standard VLM token budgets without introducing any special tokens during fine-tuning and significantly improving generative quality. In addition, to overcome the limited diversity of existing physical 3D datasets, we construct a new dataset, \textbf{\ournewdata}, which expands the object categories in prior physical 3D datasets by over \textbf{2$\times$} and includes more than 2K common real-world objects with rich physical annotations. Extensive experiments on \ournewdata{} and in-the-wild images demonstrate that \ourname\ delivers strong generative performance and robust generalization. Furthermore, simulation-based experiments in a MuJoCo-style environment validate that our sim-ready assets can be directly used for contact-rich robotic policy learning. We believe \ourname\ can substantially empower a broad range of downstream applications, especially in embodied AI and physics-based simulation.

\end{abstract}    
\section{Introduction}\label{sec:intro}

For a broad range of downstream applications in robotics, embodied AI, and interactive simulation, there is an increasing demand for high-quality physical 3D assets that can be directly executed in simulators. However, most existing 3D generation methods either focus on global 3D geometry and visual appearance~\cite{trellis,3dtopia,3dtopiaxl,wang2024llama,fang2025meshllm,ye2025shapellm}, or on part-aware generation~\cite{zhang2025bang,yang2025omnipart} that models object hierarchies and fine-grained structures. Despite their visually impressive performance, the resulting assets typically lack essential physical and articulation information—such as density, absolute scale, and joint constraints—which creates a substantial gap to real-world applications and makes these assets difficult to deploy directly in simulators or physics engines.

In parallel, a few works have started to explore the generation of articulated objects~\cite{chen2024urdformer,liu2024singapo,le2024articulate,lu2025dreamart}. Yet, due to the scarcity of large-scale high-quality annotated 3D datasets, many of these methods adopt retrieval-based paradigms: they retrieve an existing 3D model and attach plausible motions, rather than synthesizing fully novel, physically grounded assets. As a result, they provide only limited articulation information, generalize poorly to in-the-wild images, and still lack the physical attributes required for realistic simulation. While prior efforts attempt to learn deformation behavior for 3D assets~\cite{chen2025physgen3d,le2025pixie,chen2025vid2sim,jiang2025phystwin}, they often impose a homogeneous-material assumption or neglect some essential physical attributes. Even PhysXGen~\cite{cao2025physx}, which can directly generate physical 3D assets, does not yet support plug-and-play deployment in standard simulators or physics engines~\cite{todorov2012mujoco,partnetmobility}, thereby constraining its practical utility for downstream embodied AI and control tasks.

To bridge the gap between synthetic 3D assets and real downstream applications, we propose \textbf{\ourname}—the \textbf{first simulation-ready (sim-ready) physical 3D generative paradigm}. Given a single in-the-wild image, \ourname\ produces a high-quality sim-ready 3D asset, as illustrated in Fig.~\ref{fig:teaser}. Specifically, we introduce the first unified VLM-based generative model that jointly predicts geometry, articulation structure, and essential physical properties. Meanwhile, to resolve the intrinsic tension between the limited token budget of VLMs and the complexity of detailed 3D geometry, we design a new 3D representation that tokenizes geometry efficiently. This representation reduces the number of tokens by \textbf{193$\times$}, making it feasible to learn explicit geometry directly while avoiding the introduction of special tokens and new tokenizer during fine-tuning. Based on the coarse geometry generated by the VLM, we further develop a controllable flow transformer and decoder to synthesize fine-grained geometry and the corresponding URDF \& XML structure, yielding sim-ready assets that can be directly imported into standard simulators.

Additionally, to significantly enrich the diversity of existing physically grounded 3D datasets~\cite{cao2025physx}, we build a new dataset, \textbf{\ournewdata}, by collecting assets from PartNet-Mobility~\cite{partnetmobility} and carefully annotating their physical attributes. \ournewdata\ spans 47 categories and covers common real-world objects such as toilets, fans, cameras, coffee machines, and staplers, thereby substantially broadening the category coverage of physical 3D assets. Comprehensive experiments on \ournewdata, in-the-wild images, and user studies demonstrate that \ourname\ achieves strong generative quality and robust generalization compared with recent state-of-the-art methods. Furthermore, to validate executability in standard simulators and physics engines, we conduct experiments in a MuJoCo-style simulator, showing that our sim-ready assets can be directly used in robotic policy learning for contact-rich tasks, such as safe manipulation of delicate objects like eyeglasses. We believe our work opens up new possibilities and directions for future research in 3D generation, embodied AI, and robotics.

To summarize, our main contributions are: 
\begin{itemize}[nosep,nolistsep]

\item We introduce \textbf{\ourname}, the first sim-ready physical 3D generative paradigm that, given a single in-the-wild image, produces high-quality sim-ready 3D assets, thereby pushing the frontier of physically grounded 3D content creation and unlocking new possibilities for downstream applications in simulation and embodied AI.

    
    \item We propose a unified \textbf{VLM-based} generative pipeline together with a \textbf{novel physical 3D representation}. Our representation compresses geometry tokens at a high rate while preserving explicit geometric structure, and avoids introducing any special tokens during fine-tuning.
    


    \item We construct a new physically grounded 3D dataset, \textbf{\ournewdata}, which enriches the category diversity of existing physical 3D datasets by over \textbf{2$\times$}, including over 2K common real-world objects such as cameras, coffee machines, and staplers.
   
    \item Through comprehensive evaluations on \ournewdata\ and in-the-wild images, we demonstrate the strong generative quality and robust generalization of \ourname. Furthermore, we validate the feasibility of directly deploying our sim-ready assets in simulation environments, thereby empowering downstream applications such as embodied AI and robotic manipulation.
    

\end{itemize}
\section{Related Works}
\label{sec:related}

\begin{table}[b]
\vspace{-20pt}
\caption{Comparison of representative methods and their capabilities. Gen. represents the generalization of methods. It shows that our \ourname\ is the only approach that simultaneously satisfies all four criteria. } 
\vspace{-5pt}
\scriptsize
  \centering
   \setlength{\tabcolsep}{0.6mm}{
    \begin{tabular}{lccccc}
    
    \toprule[1pt]
          \textbf{Methods}&\textbf{Paradigm} & \multicolumn{1}{c}{\bf Articulate} & \multicolumn{1}{c}{\bf Physical} & \multicolumn{1}{c}{\bf Gen.} & \multicolumn{1}{c}{\bf Sim-ready} \\

          \midrule
        URDFormer \textit{.etc}~\cite{chen2024urdformer,liu2024singapo,le2024articulate}&Retrieval&\cmark & \xmark&\xmark&\xmark \\

    Trellis \textit{.etc}~\cite{trellis,3dtopia,3dtopiaxl}&Diffusion&\xmark&\xmark&\cmark&\xmark\\
    MeshLLM \textit{.etc}~\cite{wang2024llama,fang2025meshllm,ye2025shapellm}&VLM&\xmark&\xmark&\cmark&\xmark\\

       PhysXGen~\cite{cao2025physx}&Diffusion&\cmark&\cmark&\cmark&\xmark\\
     \textbf{\ourname} &VLM&\cmark&\cmark&\cmark&\cmark \\
    \bottomrule
    \end{tabular}%
  \label{tab:related}%
  }
\end{table}%

\begin{figure*}[t]
\begin{center}
\hsize=\textwidth 
\includegraphics[width=1\textwidth]{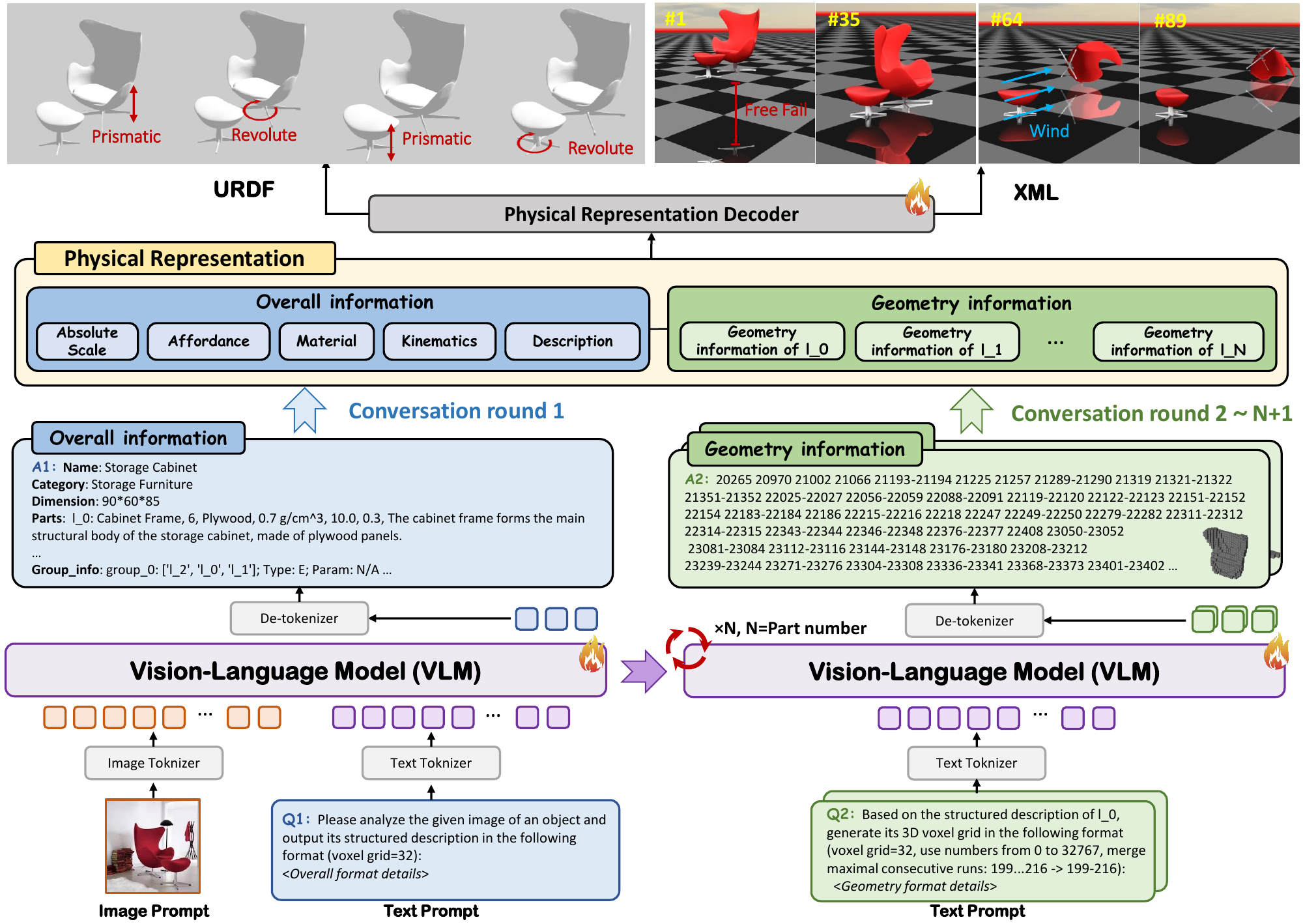}

\caption{\textbf{Overview of \ourname\ .} \ourname\ conducts a multi-round conversation to produce a physical representation that includes overall information (left) and detailed geometric information for each part (right). Decoding this representation yields high-quality, simulation-ready 3D assets with explicit physical attributes that can be directly used in downstream applications. }
\vspace{-20pt}
\label{fig:framework}
\end{center} 
\end{figure*}

\subsection{3D Generative Models}

As one of the earliest paradigms for 3D generation, generative adversarial networks (GANs) played a central role in the early stage of this field~\cite{chan2022efficient,gao2022get3d}. However, they struggle to maintain stable and robust generative performance in complex, diverse scenarios. Subsequently, DreamFusion~\cite{poole2022dreamfusion} introduced the SDS loss, which leverages the strong prior of 2D diffusion models to achieve impressive text-driven 3D generation quality. Nevertheless, optimization-based methods still suffer from the multi-face Janus problem and low optimization efficiency. Recently, feed-forward methods have become the mainstream in 3D generation due to their favorable efficiency and robustness~\cite{trellis,lgm,instantmesh,3dtopia,3dtopiaxl,difftf,difftf1,zlabor}. Beyond diffusion-based models, several works introduce autoregressive modeling into 3D generation~\cite{chen2024meshanything,siddiqui2024meshgpt}. Motivated by the strong performance of vision–language models (VLMs), recent approaches have begun to employ VLMs to generate 3D assets. To limit the token length, LLaMA-Mesh~\cite{wang2024llama} adopts a simplified mesh representation, upon which MeshLLM~\cite{fang2025meshllm} builds a part-to-assembly pipeline to further improve generative quality. Instead of using a simplified mesh representation, ShapeLLM-Omni~\cite{ye2025shapellm} adopts a 3D VQ-VAE to compress the token sequence length, but at the cost of introducing additional special tokens and a new tokenizer for geometry, which complicates the training procedure.

\begin{figure}[t]

\includegraphics[width=0.5\textwidth]{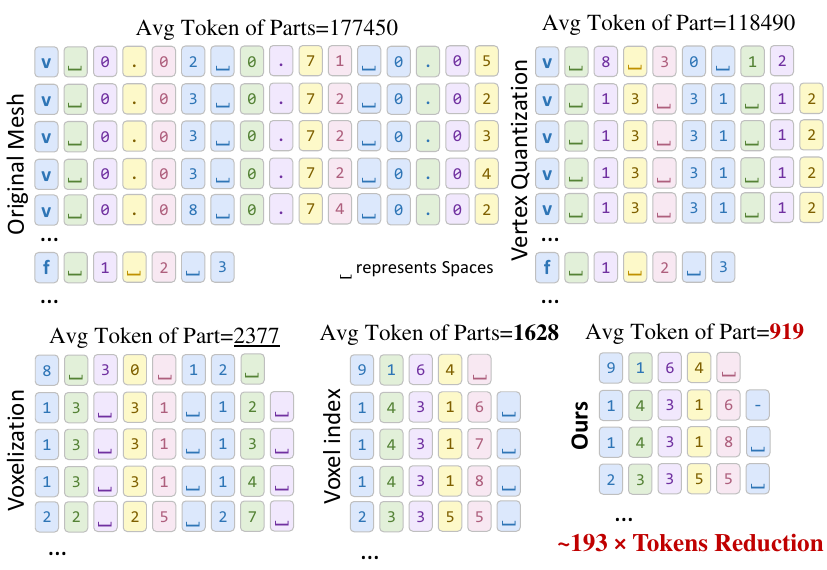}
\vspace{-15pt}
\caption{\textbf{Comparison of token counts between representations.} By adopting a voxel-based representation together with a specialized merging strategy, our method reduces the token count by \textbf{193×} compared with the original mesh format.}
\label{fig:rep}
\vspace{-10pt}
\end{figure}

In contrast to prior work, to better unlock the potential of VLMs for 3D generation, we propose a new, efficient representation that substantially compresses the token sequence while preserving explicit structural information. Moreover, our approach introduces no additional special tokens during fine-tuning, thereby avoiding both the need for large-scale task-specific pretraining datasets and the overhead of training a new tokenizer for sim-ready physical 3D generation.

\subsection{Articulated and Physical 3D Object Generation}

Articulated object generation has attracted increasing attention due to its wide range of applications. Most existing methods are retrieval-based: they first define a source library and then retrieve meshes from it to construct articulated 3D assets~\cite{chen2024urdformer,le2024articulate}. Other works adopt graph-structured representations~\cite{liu2024singapo,lei2023nap}, combining the kinematic graph of an articulated object with diffusion models to enable shape generation without texture. However, these approaches struggle to robustly generalize to novel structures, unseen categories, and complex texture. DreamArt~\cite{lu2025dreamart} instead attempts to optimize articulated 3D objects from video generation outputs, but it requires manually annotated part masks and becomes unstable when handling objects with many movable parts. URDF-Anything~\cite{li2025urdf} can directly generate URDF files. However, it relies on robust point cloud inputs and is hard to generate detailed texture for 3D assets. Although some works attempt to learn the physical deformation of 3D assets~\cite{chen2025physgen3d,le2025pixie,chen2025vid2sim,jiang2025phystwin}, they either treat all objects as homogeneous or ignore some key physical attributes. To push 3D generation toward physical realism, PhysXGen~\cite{cao2025physx} first proposes a unified framework that directly generates 3D assets with essential physical properties, including absolute dimension, density, and so on. Despite its promising performance in physical 3D generation, there remains a substantial gap between the synthesized assets and the requirements of modern physics simulators, resulting in limited direct usability in downstream tasks.

To fully realize the downstream utility of synthetic 3D assets, we introduce the first 3D generation paradigm that, from a single real-world image, produces high-quality sim-ready 3D assets equipped with explicit physical properties. We compare \ourname\ with existing approaches in Table~\ref{tab:related}, which highlights that our method is the only one that simultaneously supports articulation, physical modeling, strong generalization, and simulation-ready deployment. We believe that our approach offers a new direction for using synthetic data to empower related applications.

\begin{figure*}[t]

\includegraphics[width=1\textwidth]{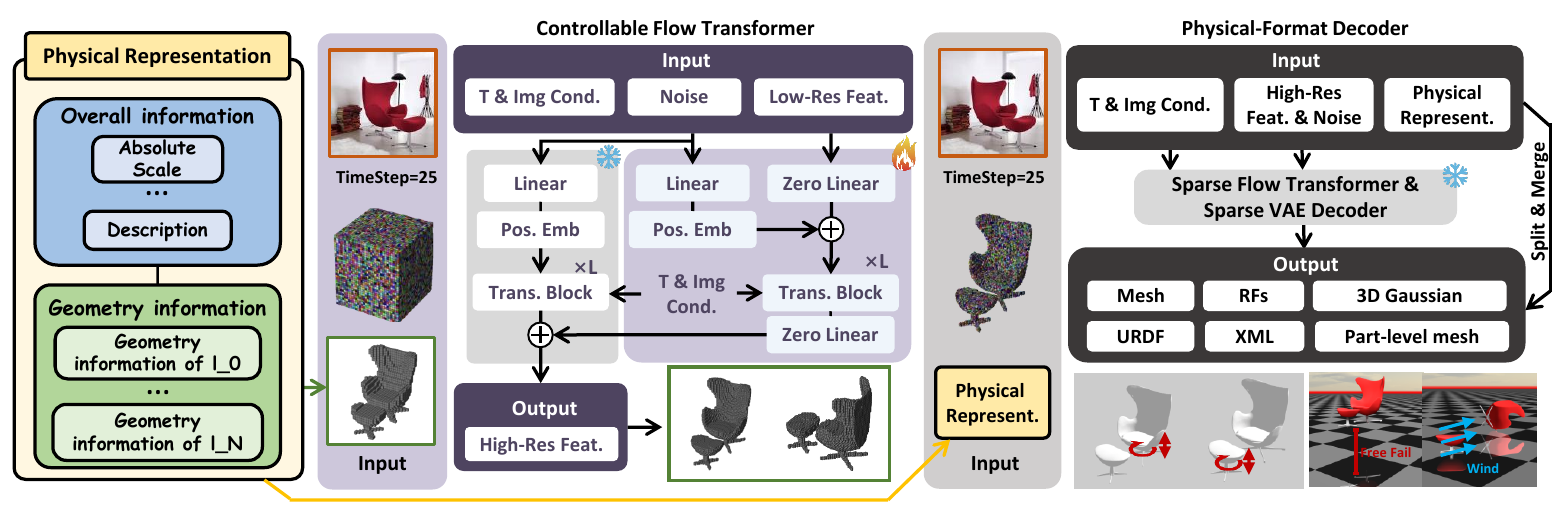}
\vspace{-20pt}
\caption{\textbf{Detailed structure of the physical representation decoder.} Given the coarse geometry, a controllable flow transformer is employed to generate fine-grained geometric information. The format decoder then combines the overall physical information and the refined geometry to produce assets in six different formats. }
\label{fig:decoder}
\vspace{-15pt}
\end{figure*}
\section{Methodology}

In this section, we present the detailed paradigm of \ourname, as illustrated in Fig.~\ref{fig:rep}. It adopts a global-to-local pipeline. Specifically, given a real-world image, \ourname\ conducts a multi-round conversation to sequentially generate the overall physical description and the geometric information of each part. To mitigate context forgetting caused by overly long prompts, we retain only the overall information when generating per-part geometry. In other words, the geometric descriptions of different parts are generated independently, conditioned solely on the shared overall information. Finally, by decoding the physical representation, \ourname\ can output simulation-ready physical 3D assets in six commonly used formats.

\subsection{Physical Representation}
Previously, to reduce the token length of raw 3D meshes in VLM-based frameworks, most 3D generation methods~\cite{wang2024llama,fang2025meshllm} adopt text-serialized representations based on vertex quantization. However, the resulting token sequences remain excessively long. Although 3D VQ-GAN~\cite{ye2025shapellm} can further compress geometric tokens, it requires introducing additional special tokens and a customized tokenizer during fine-tuning, which complicates training and deployment.

To address these limitations, we propose a new 3D representation that substantially reduces token length while preserving explicit geometric structure, without introducing any additional tokenizer. Motivated by the impressive trade-off between fidelity and efficiency of voxel-based representations~\cite{trellis}, we build our representation on voxels. Directly encoding high-resolution voxels, however, still yields an unaffordable number of tokens for VLMs, even after mapping geometry to a compressed space. We therefore adopt a coarse-to-fine strategy for geometry modeling: the VLM operates on a $32^3$ voxel grid to capture coarse geometry, while a downstream decoder refines this coarse shape into high-fidelity geometry. In this way, we retain the explicit structural advantages of 3D voxels while avoiding excessive token consumption. As shown in Fig.~\ref{fig:rep}, converting meshes to coarse voxels alone reduces the number of tokens by $74\times$. To further eliminate redundancy in sparse voxel data, we linearize the $32^3$ grid into indices from 0 to $32^3-1$ and serialize only occupied voxels. Finally, by merging neighboring occupied indices and connecting continuous ranges with a hyphen $-$, we achieve an even higher token compression rate (\textbf{193$\times$}) while maintaining explicit geometric structure.

For overall information, we adopt a tree-structured, VLM-friendly representation following~\cite{cao2025physx}. Compared with standard URDF files, our JSON-style format provides richer physical attributes and textual descriptions, thereby facilitating understanding and reasoning by VLMs. Moreover, to maintain consistency between kinematic structure and geometry, we convert key kinematic parameters into the voxel space, including direction of motion, axis location, motion range, and related articulation properties.

\begin{figure*}[t]
\begin{center}
\hsize=\textwidth 
\includegraphics[width=1\textwidth]{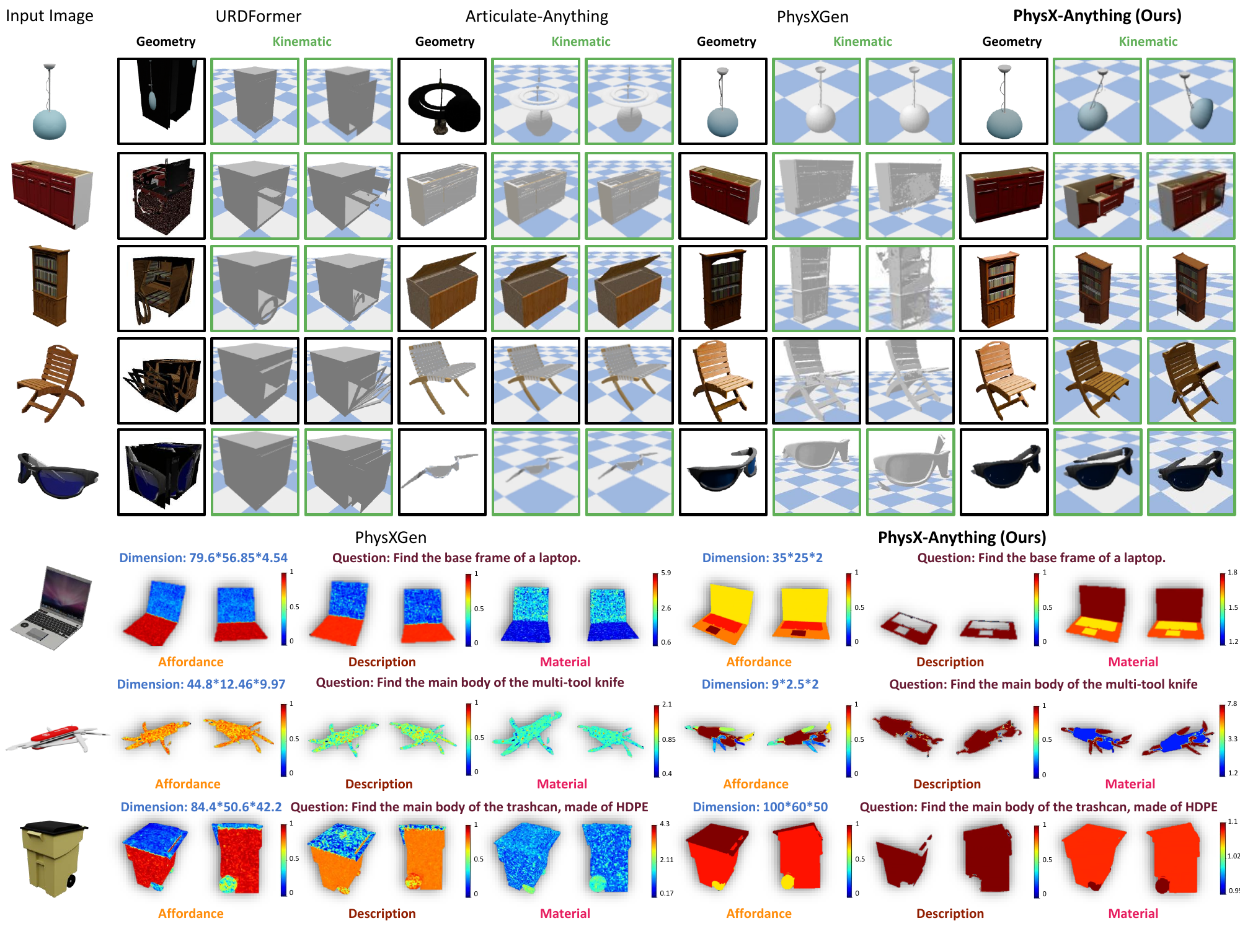}
\vspace{-30pt}
\caption{\textbf{Qualitative results on the test set of \ournewdata.} Compared with other methods, \ourname\ generates high-quality, sim-ready physical 3D assets with more faithful geometry, articulation, and physical attributes.}
\vspace{-15pt}
\label{fig:eval}
\end{center} 
\end{figure*}

\begin{table*}[t]

\caption{\label{table:eval}\textbf{Quantitative comparison with other methods on \ournewdata.} \ourname\ consistently outperforms all SOTA methods across all metrics, with especially large gains on physical properties.}
\vspace{-5pt}
\scriptsize
  \centering
   \resizebox{1\textwidth}{!}{%

    \begin{tabular}{@{}l|ccc|ccccc|@{}}
    \toprule
          \multirow{2}{*}{\bf Methods}& \multicolumn{3}{c|}{\bf Geometry} &  \multicolumn{5}{c|}{\bf Physical Attributes} \\
          
          & \multicolumn{1}{c}{\bf PSNR $\uparrow$} & \multicolumn{1}{c}{\bf CD $\downarrow$} & \multicolumn{1}{c|}{\bf F-score $\uparrow$} &    \multirow{1}{*}{\bf \textcolor{color2}{Absolute scale} $\downarrow$} &  \multirow{1}{*}{\bf \textcolor{color3}{Material} $\uparrow$} & \multirow{1}{*}{\bf \textcolor{color1}{Affordance} $\uparrow$} & {\bf \textcolor{color4}{Kinematic parameters (VLM)} $\uparrow$} &  \multirow{1}{*}{\bf \textcolor{color5}{Description} $\uparrow$} \\
          \midrule
    URDFormer~\cite{chen2024urdformer} &   7.97    &   48.44    &  43.81     & --&--&--&0.31&-- \\
    Articulate-Anything~\cite{le2024articulate} &    16.90   &  17.01     &   67.35    &  --&--&--&0.65&--\\
    PhysXGen~\cite{cao2025physx}&20.33 &14.55&76.3&43.44&6.29&9.75&0.71&12.89\\
    \midrule
    \bf \ourname\ (Ours) &   \bf 20.35    &   \bf 14.43    &   \bf 77.50    & \bf 0.30 & \bf 17.52&\bf 14.28&\bf 0.83&\bf 19.36  \\
     \bottomrule
    \end{tabular}%

}
\vspace{-10pt}
\end{table*}

\begin{figure*}[t]
\begin{center}
\hsize=\textwidth 
\includegraphics[width=1\textwidth]{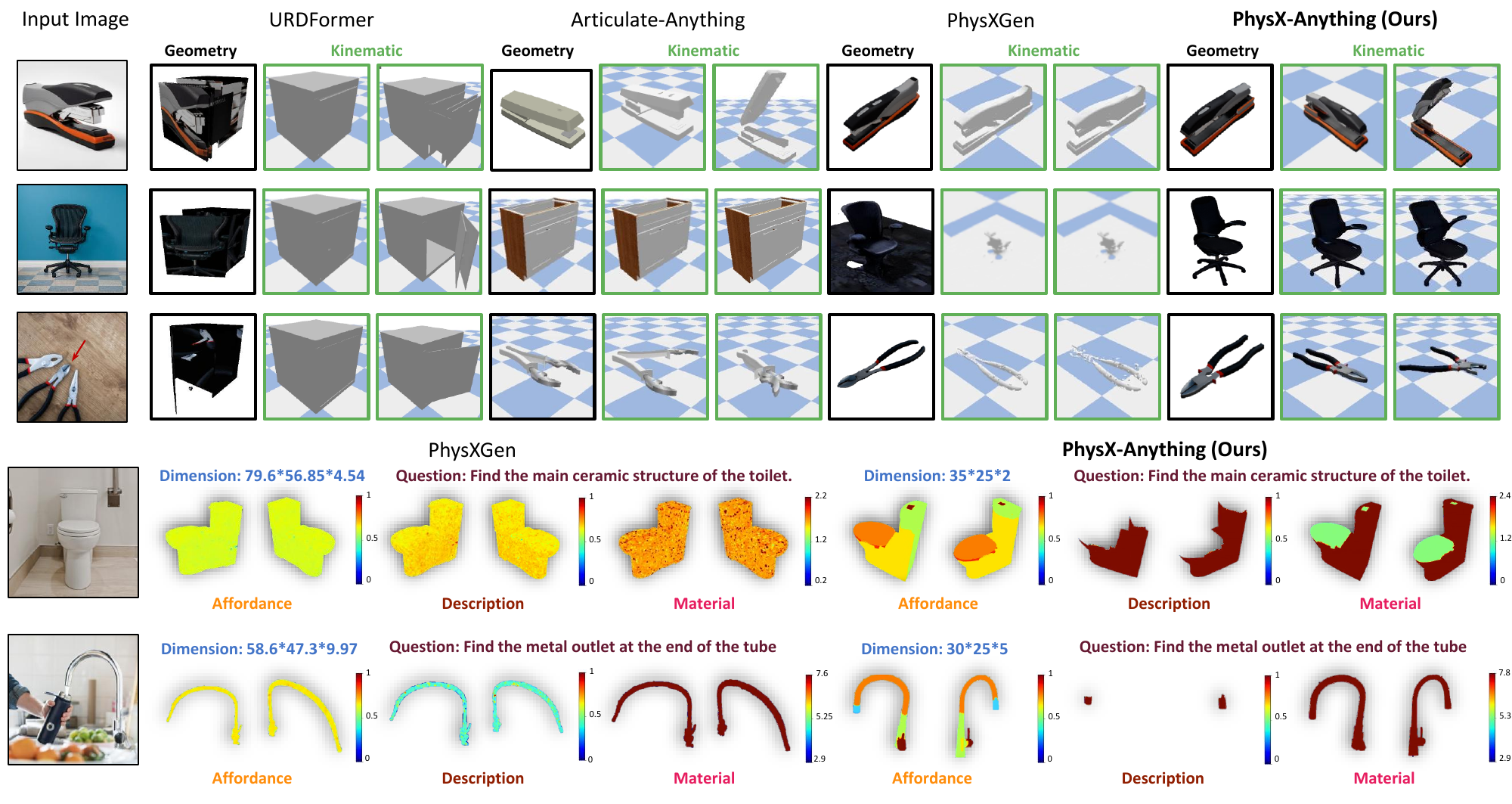}
 \vspace{-20pt}
\caption{\textbf{Qualitative results on in-the-wild images.} Given a single real-world image as input, \ourname\ produces high-quality sim-ready 3D assets with realistic geometry, articulation, and physical attributes across diverse object categories. Moreover, the results highlight the robust generalization of \ourname. }
 \vspace{-10pt}
\label{fig:wild}
\end{center} 
\end{figure*}

\begin{table*}[t]
\caption{\label{table:user} \textbf{User studies on in-the-wild evaluation.} User preference results on in-the-wild cases show that \ourname\ significantly outperforms other methods, achieving a clear margin of improvement in geometry quality and physical plausibility.}

\scriptsize
  \centering
   \setlength{\tabcolsep}{3mm}{
  \begin{tabular}{@{}l|c|ccccc|@{}}
      \toprule

     \multirow{2}{*}{\bf Methods} & \multirow{2}{*}{\bf Geometry (Human) $\uparrow$} & \multicolumn{5}{c|}{\bf Physical Attributes (Human)} \\
     
     & &  \multirow{1}{*}{\bf \textcolor{color2}{Absolute scale} $\uparrow$} &  \multirow{1}{*}{\bf \textcolor{color3}{Material} $\uparrow$} & \multirow{1}{*}{\bf \textcolor{color1}{Affordance} $\uparrow$} & \multicolumn{1}{c}{\bf \textcolor{color4}{Kinematic parameters} $\uparrow$} &  \multirow{1}{*}{\bf \textcolor{color5}{Description} $\uparrow$}
      \\
      
    \midrule

     URDFormer~\cite{chen2024urdformer} &0.21&--&--&--&0.23&--\\
    Articulate-Anything~\cite{le2024articulate}&0.53&--&--&--&0.37&--\\
    PhysXGen~\cite{cao2025physx}&0.61&0.48&0.43&0.34&0.32&0.33\\
    \midrule
    \bf \ourname\ (Ours) & \bf 0.98&\bf 0.95&\bf 0.84 &\bf 0.94&\bf 0.98&\bf 0.96 \\

    \bottomrule
    \end{tabular}
    }
\vspace{-10pt}
\end{table*}

\begin{table}[t]
\caption{\label{table:wild} \textbf{In-the-wild VLM-based evaluation.} Quantitative results from GPT-5 also confirm the strong generative performance of \ourname\ in terms of geometry and articulation.}

\scriptsize
  \centering
   \setlength{\tabcolsep}{1.5mm}{
    
    \begin{tabular}{@{}l|c|c|@{}}
\toprule
          \multirow{1}{*}{\bf Methods}& \multicolumn{1}{c|}{\bf Geometry (VLM) $\uparrow$}   &\multirow{1}{*}{\bf \textcolor{color4}{Kinematic parameters (VLM)} $\uparrow$} \\

          \midrule
    URDFormer~\cite{chen2024urdformer} &  0.29     &        0.31        \\
    Articulate-Anything~\cite{le2024articulate} &  0.61     &       0.64       \\
    PhysXGen~\cite{cao2025physx}&0.65&       0.61         \\
    \midrule
    \bf \ourname\ (Ours) &  \bf 0.94    &       \bf 0.94          \\
    \bottomrule
    \end{tabular}%
    
  \label{tab:addlabel}%
}
\vspace{-10pt}
\end{table}

\begin{figure*}[t]
\begin{center}
\hsize=\textwidth 
\includegraphics[width=1\textwidth]{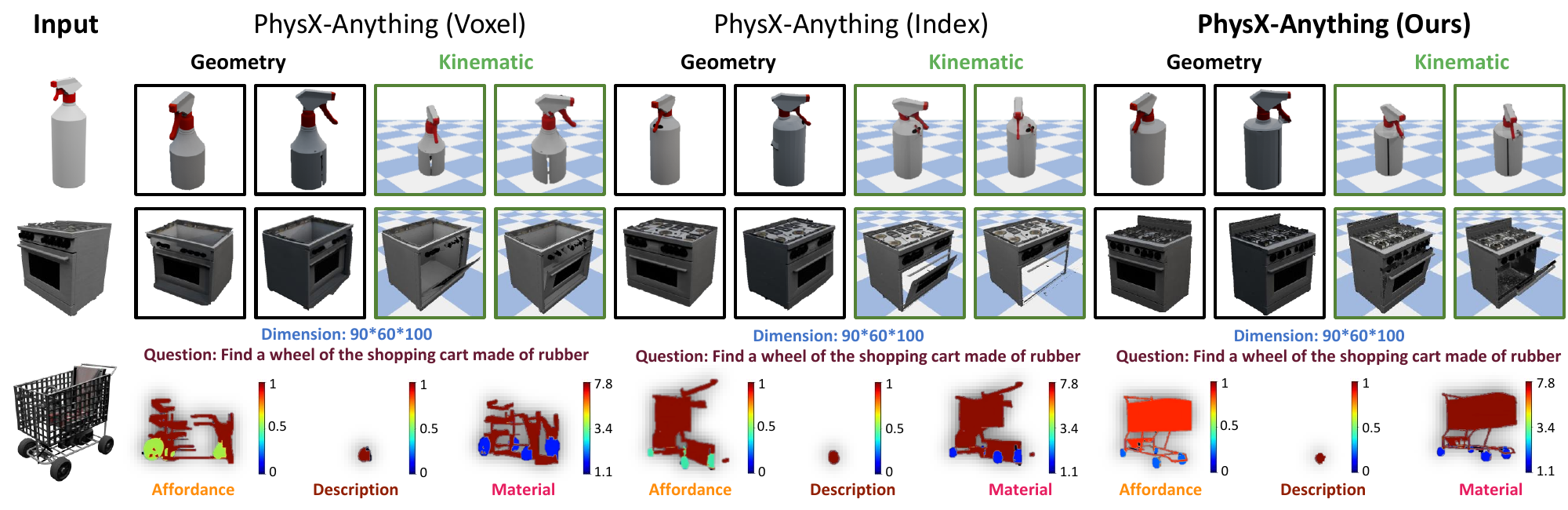}
\caption{\label{fig:ablation} \textbf{Ablation studies on different representation.} We compare the generative performance of different 3D representations, which validates both the effectiveness and efficiency of our proposed representation.}

\includegraphics[width=1\textwidth]{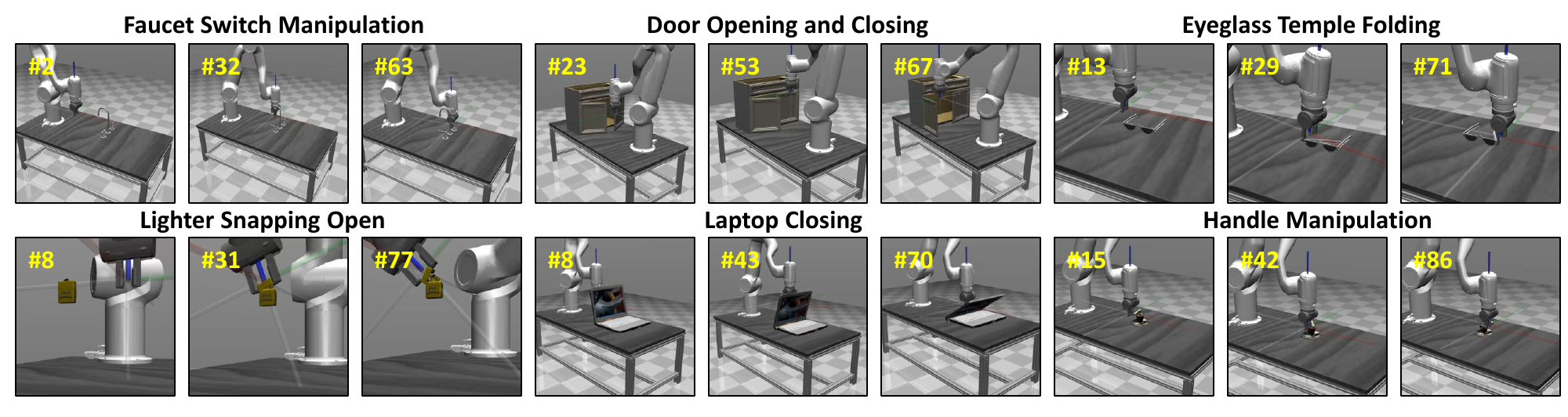}
\vspace{-20pt}
\caption{\label{fig:sim} \textbf{Robot Manipulation on Generated sim-reaady 3D assets of \ourname.} The results show that our generated sim-ready assets exhibit highly physically plausible behavior and accurate geometric structure across diverse tasks, thereby providing a new direction for robotics policy learning.}
\vspace{-10pt}
\end{center} 
\end{figure*}
\begin{table*}[t]
\vspace{-5pt}
\caption{\label{table:ablation} \textbf{Comparison with different representations.} Quantitative results across different 3D representations clearly demonstrate the superiority of our proposed representation in both geometric fidelity and physical attributes.}

\scriptsize
  \centering
   \setlength{\tabcolsep}{2mm}{
  \begin{tabular}{@{}l|ccc|ccccc|@{}}
      \toprule

     \multirow{2}{*}{\bf Methods} & \multicolumn{3}{c|}{\bf Geometry} & \multicolumn{5}{c|}{\bf Physical Attributes} \\
     
     & \multicolumn{1}{c}{\bf PSNR $\uparrow$} & \multicolumn{1}{c}{\bf CD  $\downarrow$}& \multicolumn{1}{c|}{\bf F-Score $\uparrow$}& \multirow{1}{*}{\bf \textcolor{color2}{Absolute scale} $\downarrow$} &  \multirow{1}{*}{\bf \textcolor{color3}{Material} $\uparrow$} & \multirow{1}{*}{\bf \textcolor{color1}{Affordance} $\uparrow$} & \multicolumn{1}{c}{\bf \textcolor{color4}{Kinematic parameters (VLM)} $\uparrow$} &  \multirow{1}{*}{\bf \textcolor{color5}{Description} $\uparrow$}
      \\
      
    \midrule

    
    \ourname-Voxel &    16.96    &    17.81    &    63.10 &0.40&12.32&11.63&0.39&17.38\\
    \ourname-Index &    18.21    &    16.27  &    68.70 &0.30&13.35&12.04&0.76&17.97\\
    \bf \ourname\ (Ours) &   \bf 20.35    &   \bf 14.43    &   \bf 77.50    & \bf 0.30 & \bf 17.52&\bf 14.28&\bf 0.94&\bf 19.36  \\

    \bottomrule
    \end{tabular}
    }
\vspace{-9pt}
\end{table*}


\subsection{VLM \& Physical Representation Decoder}
Building on the above representation for physical 3D assets, we adopt Qwen2.5~\cite{bai2025qwen2} as our foundation model and fine-tune the VLM on our physical 3D datasets. Through a tailored multi-round dialogue, \ourname\ then generates both high-quality global descriptions (overall physical and structural properties) and local information (part-level geometry). To obtain more detailed geometry, we further design a controllable flow transformer inspired by ControlNet~\cite{zhang2023adding}. Built on the flow transformer architecture~\cite{trellis}, we introduce a transformer-based control module that takes the coarse voxel representation as guidance for the diffusion model, thereby steering the synthesis of fine-grained voxel geometry. Thus, the training objective of the controllable flow transformer is formulated as:
\begin{equation}
\mathcal{L}_{\text{geo}}
= \mathbb{E}_{t, x_0, \epsilon, c, \mathbf{V}^{\text{low}}}
\bigl[
\bigl\|
f_\theta(x_t, c, \mathbf{V}^{\text{low}}, t) - (\epsilon - x_0)
\bigr\|_2^2
\bigr]~,
\end{equation}
where $\mathbf{V}^{\text{low}}$, $x_0$, $\epsilon$, $c$, $t$, and $f_\theta$ denote the coarse voxel representation, fine-grained voxel target, Gaussian noise, image condition, time step, and the controllable flow transformer parameterized by $\theta$, respectively. The noisy sample $x_t$ is obtained by interpolating between $x_0$ and $\epsilon$, \textit{i.e.},$x_t = (1 - t)x_0 + t\epsilon~.$

Given the fine-grained voxel representation, we adopt a pre-trained structured latent diffusion model~\cite{trellis} to generate 3D assets, including mesh surfaces, radiance fields, and 3D Gaussians. We then apply a nearest-neighbor algorithm to segment the reconstructed mesh into part-level components, conditioned on the voxel assignments. Finally, by combining the global structural information with the fine-grained voxel geometry, \ourname\ can generate URDF, XML, and part-level meshes for sim-ready physical 3D generation.

\section{Experiments}

In this section, we present experimental results on \ournewdata\ and in-the-wild images. More details are provided in the supplementary material.

\subsection{Evaluation on \ournewdata}

We compare \ourname\ with the most related state-of-the-art methods, URDFormer~\cite{chen2024urdformer}, Articulate-Anything~\cite{le2024articulate}, and PhysXGen~\cite{cao2025physx}. As shown in Table~\ref{table:eval}, \ourname\ consistently achieves the best performance across both geometric and physical metrics. Benefiting from the strong VLM prior, \ourname\ yields a dramatic improvement in \textbf{\textcolor{color2}{absolute scale}} (reducing the error from 43.44 to 0.30, i.e., over \textbf{99\%} relative improvement compared with PhysXGen). Moreover, since VLMs are inherently text-friendly, \ourname\ also attains the highest scores on \textbf{\textcolor{color5}{description}}, indicating that our method not only produces physically plausible properties but also generates coherent, part-level textual descriptions that reflect a strong understanding of object structure and function.

Beyond the quantitative comparison, we further present qualitative results in Fig.~\ref{fig:eval}. It clearly highlight the superiority of \ourname\ in terms of generalization, especially when compared with retrieval-based methods~\cite{chen2024urdformer,le2024articulate}. Leveraging the powerful VLM prior and efficient representation, \ourname\ also produces significantly more plausible physical attributes than PhysXGen~\cite{cao2025physx}.

\subsection{In-the-Wild Evaluation}
\noindent\textbf{VLM-based evaluation.}
To evaluate generalization in real-world scenarios, we collect approximately 100 in-the-wild images from the Internet using category keywords. These real-world images cover the most common everyday object categories. To avoid unreliable VLM judgments on specific physical properties, we focus the VLM-based evaluation on geometry and articulation quality. As reported in Table~\ref{table:wild}, \ourname\ achieves substantially higher scores than all competing methods on both geometry (VLM) and kinematic parameters (VLM), indicating markedly better generalization to real-life inputs.

\noindent\textbf{User studies on real-life images.} 
To complement the in-the-wild VLM evaluations on physical attributes, we conduct user studies, as summarized in Table~\ref{table:user}. Each volunteer rates the generated results on a 0 to 5 scale, considering both geometry and all physical attributes. In total, we collect 1,568 valid scores from 14 volunteers and normalize the scores. The results show that the outputs of \ourname\ align much better with human preferences than those of other methods, confirming its robust generative performance in both geometry and physical properties. The visualizations in Figure~\ref{fig:wild} on real-life scenarios further highlight the superiority of \ourname against other methods, showing more accurate geometry, articulation, and physical attributes across diverse and challenging in-the-wild cases.

\subsection{Ablation Studies}
To analyze the effectiveness of our representation, we conduct ablation studies over different designs, as illustrated in Fig.~\ref{fig:rep}. Note that the original mesh and vertex-quantization representations require an excessively large number of tokens, making end-to-end training infeasible due to out-of-memory issues. Therefore, we focus our comparison on the remaining three compact representations. As shown in Table~\ref{table:ablation}, as the token compression ratio increases, \ourname\ is able to capture complete and detailed geometry even for complex structures, whereas alternative representations are constrained by the token budget and suffer noticeable degradation. The qualitative results in Fig.~\ref{fig:ablation} further show that our \ourname\ produces more robust results for geometrically challenging objects.

\subsection{Robotic Policy Learning in Simulation}

To validate the potential of our approach for supporting downstream tasks, we conduct experiments in a MuJoCo-style simulator~\cite{Zhou_robopal_A_Simulation_2024}, as illustrated in Fig.~\ref{fig:sim}. Our generated simulation-ready 3D assets—including faucets, cabinets, lighters, eyeglasses, and other everyday objects—can be directly imported into the simulator and used for contact-rich robotics policy learning. This experiment not only demonstrates the physically plausible behavior and accurate geometry of our generated assets, but also highlights their strong potential to enable and inspire a wide range of downstream robotics and embodied AI applications. 
\section{Conclusion}

In this paper, we aim to fully unlock the potential of synthesized 3D assets in real-world applications by introducing \ourname, the first sim-ready physical 3D generative paradigm. Through a unified VLM-based pipeline and a tailored 3D representation, \ourname\ achieves substantial token compression (over 193$\times$) while preserving explicit geometric structure, enabling efficient and scalable physical 3D generation. In addition, to enrich the diversity of existing physical 3D datasets, we construct \ournewdata\ by carefully collecting and annotating common real-world objects with rich physical attributes. It includes 47 the most common real-life categories with detailed physical attributes. Comprehensive experiments on \ournewdata\ and in-the-wild scenarios demonstrate the strong performance and robust generalization of \ourname\ in sim-ready physical 3D generation. Furthermore, simulation-based experiments highlight its potential for downstream robotic policy learning. We believe \ourname\ will spur new research directions across 3D vision, embodied AI and robotics. 


{
    \small
    \bibliographystyle{ieeenat_fullname}
    \bibliography{main}
}


\end{document}